\documentclass[conference]{IEEEtran}

\ifCLASSINFOpdf
\else
\fi

\usepackage{amssymb}
\usepackage{latexsym}
\usepackage{caption}
\usepackage{url}
\usepackage{subcaption}
\usepackage{graphicx}
\usepackage{textcomp}
\usepackage{enumitem}
\usepackage{algorithm,algorithmic}

\begin{document}
%
\title{Automated Parking Space Detection \\Using Convolutional Neural Networks}

\author{\IEEEauthorblockN{Julien Nyambal}
\IEEEauthorblockA{School of Computer Science and\\Applied Mathematics\\
University of the Witwatersrand\\
Johannesburg, South Africa\\
Email: julien.nyambal1@students.wits.ac.za}
\and
\IEEEauthorblockN{Richard Klein}
\IEEEauthorblockA{School of Computer Science and\\Applied Mathematics\\
University of the Witwatersrand\\
Johannesburg, South Africa\\
Email: richard.klein@wits.ac.za}
}


%


\maketitle

\begin{abstract}
Finding a parking space nowadays becomes an issue that is not to be neglected, it consumes time and energy. We have used computer vision techniques to infer the state of the parking lot given the data collected from the University of The Witwatersrand. This paper presents an approach for a real-time parking space classification based on Convolutional Neural Networks (CNN) using Caffe and Nvidia DiGITS framework. The training process has been done using DiGITS and the output is a caffemodel used for predictions to detect vacant and occupied parking spots. The system checks a defined area whether a parking spot (bounding boxes defined at initialization of the system) is containing a car or not (occupied or vacant). Those bounding box coordinates are saved from a frame of the video of the parking lot in a JSON format, to be later used by the system for sequential prediction on each parking spot. \newline
The system has been trained using the LeNet network with the Nesterov Accelerated Gradient as solver and the AlexNet network with the Stochastic Gradient Descent as solver. We were able to get an accuracy on the validation set of 99\% for both networks. The accuracy on a foreign dataset(PKLot) returned as well 99\%. Those are experimental results based on the training set shows how robust the system can be when the prediction has to take place in a different parking space.
\end{abstract}


%
\IEEEpeerreviewmaketitle

\section{Introduction}
As the population grows, the number of private vehicles increases as well. But the number of parking spaces most of the time remains. Sometimes, and most of the time, there are vacant spots, but the drivers do not have any information about them. It could be, either the free spot is far from them, or it is hidden by some other cars or any other objects big enough to hide the spot. In the past, and maybe at some places now, parking spaces are managed by some persons in the parking lot who might not have the total view of the next available parking space. Sometimes the driver him/herself has to check for a vacant space by circling in the parking lot, and another driver will come and many losses are generated: time, fuel, and maybe temper. Some researchers came up with different parking space detection algorithms using gadgets like video cameras or sensors to detect a vacant spot when needed. Those algorithms combined gave birth to automated parking detection.

The target of the system is mostly open areas like shopping mall parking lots or university parking lots. Implementing a sensor-based approach to automating the parking spot detection will require paperwork, time will be wasted and more trouble to the users when trying to park their vehicles while installing those devices on the ground. On the other hand, those parking lots are monitored by video cameras for security purposes, which is the case in most shopping malls as well. Video-based detection for detection is being used in many areas like obstacle detection, human detection \cite{Obs_Detection}. 

Convolutional Neural Networks architecture (CNN) is similar to the human neural network build with synapses (weights) and neurons. From this point of view, complex tasks can be learned through the network. This uses CNN with pre-existing architectures to detect in real-time the vacancy of a parking spot.  

\section{Related Work}

A sensor-based approach can be used to detect a moving object entering the parking lot. The size of the object and the time it takes to cross the gate where the sensor is located infers the occupation of a spot by the car \cite{Lee2008}. Due to the cost linked to the installation and maintenance of sensors, their use is broad but restricted when budget is an issue. Another big issue of the sensors, they cannot be implemented outdoors, because there is no roof to install a sensor.

The approach in \cite{Tschentscher} presents a video-based system for parking space classification that has been tested and resulted in a success when it has been also working on another parking lot without re-training the model. To solve the problem of occupancy of a parking spot, color histograms and Difference-of-Gaussian (DoG) have been used for feature extraction, and three machine learning algorithms have used for classifying between a free or occupied parking spot. Those three learning algorithms are: K-Nearest Neighbor (k-NN), Linear Discriminant Analysis (LDA) and the Support Vector Machines (SVM).

To avoid errors during the classification process, temporal integration exponential smoothing has been applied to the classification results. DoG and color histograms (features extractors) have been combined to the three machine learning algorithms in a one-to-one relationship, to produce the best performing combination of features extractors and machine learning algorithms used. The system has been tested with a success rate of more than 92\% on an unseen parking lot\cite{Tschentscher}.

Wu et al. \cite{Wu} proposed a method for detecting available parking space using both Support Vector Machines and a Markov Random Field framework. The input is a video, and from the images, the system distinguishes between empty spaces from occupied spaces in a parking lot. The system is trained to recognize empty parking spaces by applying a machine learning algorithm rather than segmenting the image of the vehicle from the parking space. The system is composed of 4 parts: pre-processing, ground model feature extraction, multi-class Support Vector Machines recognition, and Markov Random Field based correction. Features are extracted from the color histogram, 50 features are selected after dimension reduction. Support vector machines are used to classify the occupancy of a parking spot based on the patches of the parking lot manually determined. Markov Random Field (MRF) is used to improve the performance done by the Support Vector Machines. The results of the experiments show that Support Vector Machines used without Markov Random Field produce 84.35\% using a single space for detection and 85.57\% using 3 spots. When using Markov Random Field, the accuracy climbs to 93.53\%. The accuracy increases with the training sample, which shows the strength of the algorithm used.

A robust dataset of images of parking spaces under different weather conditions can be used to better generalize classification of the state of a spot \cite{deAlmeida20154937}. The authors use the Local Binary Pattern (LPB) and the Local Phase Quantization (LPQ) textual descriptors for feature extraction. For their classifier, they combined the Support Vector Machines with some variants of the LPQ and the LPB which are the LBP Rotation Invariant, LPQ with Gaussian window, and the LPQ gaussian derivative quadrature filter pair to achieve a classification rate of 99.64\%.

Amato et al. \cite{7543901} present an approach to detecting parking spots using deep learning techniques. The authors trained their classifier on two datasets: PKLot \cite{deAlmeida20154937}, and their own dataset from the National Research Council in Pira (CNRPark). Two network configurations have been used: mAlexNet: is a mini version of the AlexNet network and mLeNet: is the mini version of the LeNet network.

Those mini-networks are modified well-known CNN architectures (AlexNet and LeNet respectively) to accelerate the process of classification, as the original ones are very expensive in time when resources are not equivalent to the big size of the network. This method produces an accuracy greater than 90\% inter and intra-datasets.

\section{The dataset}\label{thedataset}
The dataset is made of 782 images of parking lots. Those parking spots have been extracted from 2 parking lots of the University of the Witwatersrand.

\subsection{Data Collection}
The data collection, using a GoPro Hero 4 Black camera, took a working week (5 days), from 6 am to approximately 8 am at the rate of one frame per minute. During that period of time, the parking gets filled actively and rapid change of illumination is captured. The camera was set to take pictures at 5 MegaPixels which is in a resolution of 2560 x 1920 pixels. Due to the ``\textit{Wide}'' setting of the GoPro, severe radial distortion appears on all the frames. It is of type Barrel due to the fisheye lens of the camera \cite{351044}, which affects the straight lines of the parking spots as they tend to be curved. The position of the camera and the choice of the buildings have been chosen in such a way that a car from a parking spot does not hide more than 60\% of the next car around.

\begin{figure}[!t]
	\centering
	\begin{subfigure}{0.5\textwidth}
		\centering
		\includegraphics[width=0.5\linewidth]{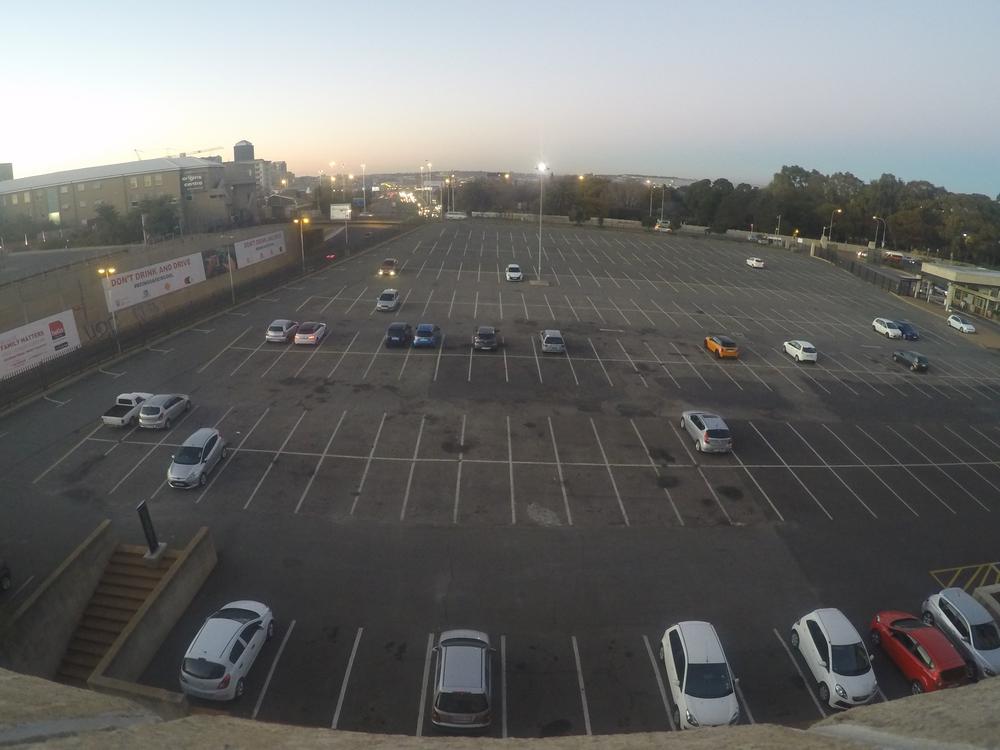}
		\caption{Original image with radial distortion of type Barrel.}
		\label{fig:sub1}
	\end{subfigure}
	\hfill
	\begin{subfigure}{0.5\textwidth}
		\centering
		\includegraphics[width=0.5\linewidth]{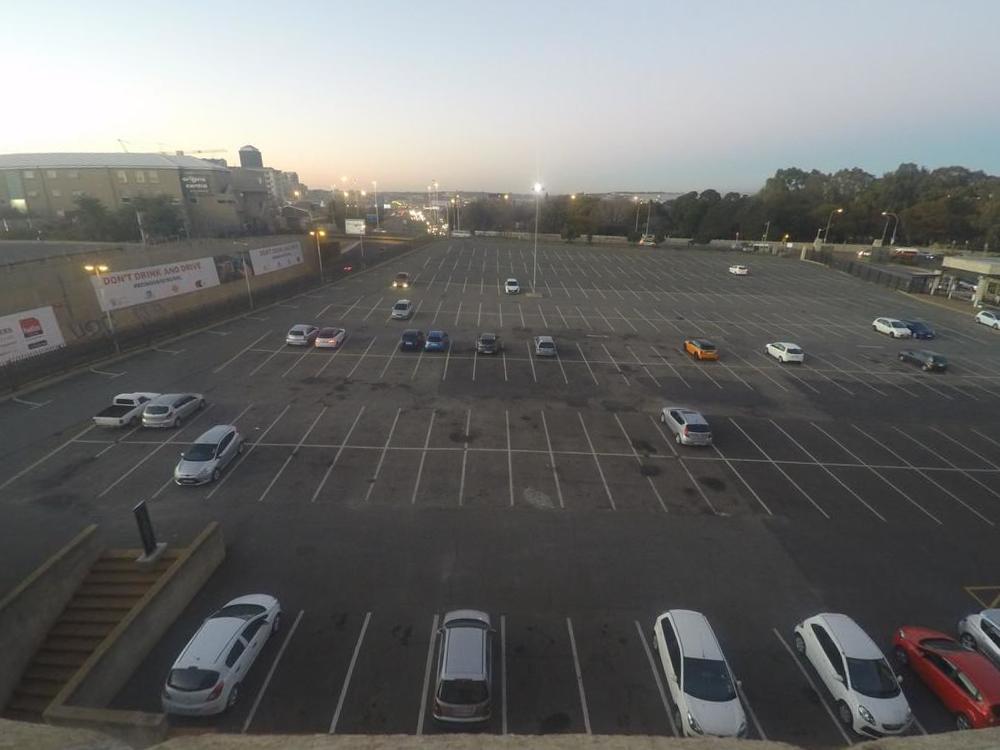}
		\caption{Image with radial distortion corrected.}
		\label{fig:sub2}
	\end{subfigure}
	\caption{Correction of radial distortion (Chamber of Mines,University of the Witwatersrand).}
	\label{fig:test}
\end{figure}

The radial distortion (Barel) has been manually fixed using distort tools and the size of the images manually fixed using convert. Distort and convert are both Unix utilities \cite{ImageMagic}. Each frame has been resized from 2560 x 1920 pixels to 1000 x 1000 pixels. \figurename{\ref{fig:sub1}} is the original frame captured by the camera that is distorted due to the fisheye lens of the camera. \figurename{\ref{fig:sub2}} is the result of all the manipulations on the original images coming directly from the camera to correct the distortion.

\subsection{Labeling}

The process used is to crop from each image of a parking lot each individual parking spot so that a label (occupied or vacant) can be manually assigned to it. To do so, the cartesian coordinates of each spot stored in an XML file generated by a labeling tool will help to extract them from the image.

\begin{figure}[!t]
	\centering
	\begin{subfigure}{0.4\textwidth}
		\centering
		\includegraphics[width=0.5\linewidth]{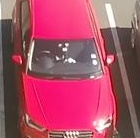}
		\caption{Occupied parking spot.}
		\label{fig:occupied}
	\end{subfigure}
	\hfill
	\begin{subfigure}{0.4\textwidth}
		\centering
		\includegraphics[width=0.5\linewidth]{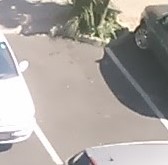}
		\caption{Vacant parking spot.}
		\label{fig:vacant}
	\end{subfigure}
	\caption{Samples of vacant and occupied parking spot.}
	\label{fig:vacancy}
\end{figure}

That tool extracts the patches of all the manually selected parking spots, using the XML file generated from a single image. Since the parking lot does not move, the same XML file has been reused on the other images to get the multiple patches \cite{cite_key, Russell2008}. \figurename{\ref{fig:occupied}} and \figurename{\ref{fig:vacant}} represent respectively the cropped image of an occupied and vacant parking spot. 

\begin{figure}[!t]
	\centering
	\includegraphics[scale=.5]{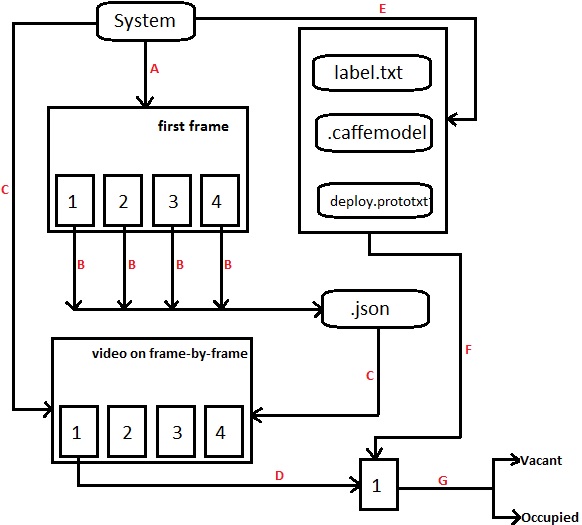}
	\caption{System Processing Flow.}
	\label{fig:process}
\end{figure}

After the labeling, vacant and occupied parking spots patches have been separated into different folders, vacant and occupied folders respectively. The raw dataset contains 782 images of parking lots.

\section{Deep Learning with Caffe and Nvidia DIGITS}

\subsection{The System}\label{system}

The system built works on the fact that the images of the parking spot will represent a fixed parking lot. 

The system analyzes the parking lot after every fifth frame. A frame represents a state of the parking spot at a precise moment. The system works on live feed or on a recorded video. \figurename{\ref{fig:process}} represents the flow of the data from raw images from video frames to take a decision on the vacancy of a parking spot.

On \figurename{\ref{fig:process}}, first frame and video on frame-by-frame represent a state of the parking lot and ``1'', ``2'', ``3'', ``4'' represent the parking spots on each frame.

\begin{enumerate}[label=\Alph*-]
	\item The System launches the frame required to define the parking spots,
	\item The coordinates of the parking spots are stored in a JSON file (.json), will get generated if it does not exist,
	\item The System loads the video and the JSON file (.json),
	\item \figurename{\ref{fig:process}} corresponds to one iteration of the loop in the number of parking spots. The first iteration will crop the spot one from the frame n of the loaded video,
	\item The System loads the labels file, the caffemodel, and the deploy file corresponding to the caffemodel,
	\item Those three arguments on the system are applied to the cropped image of the parking lot, save on the disk. In \figurename{\ref{fig:process}}, the cropped image is the parking spot "1",
	\item The result of the prediction from the System with the arguments applied on the frame n. A red or green rectangle corresponding respectively to either an occupied or vacant parking spot will surround the cropped image after the classification.
\end{enumerate} 

\subsection{Configuration}

Nvidia Deep Learning package DIGITS, which a web server that manages the activities of the GPU (Nvidia only) to efficiently perform the classification tasks. The background heavy tasks are done with Caffe. was used for both the training and testing phases. It has been used for training and testing. 

The CPU could also be used for training and testing but the performances will be relatively slower if a single CPU is to be used.

\subsubsection{Hardware and software}

The implementation of the system has been done on a desktop with an Intel i5 processor at 3.20GHz with 4 cores. The GPU used is a Nvidia GeForce GTX 750 Ti with 2GB video RAM.

The system is trained under Ubuntu 14.04 with Nvidia DIGITS 3, using cuDNN \cite{NVIDIAcu83:online} which has been shown to accelerate CNN training and testing processes \cite{DBLP:journals/corr/ChetlurWVCTCS14}. 

\subsubsection{Different Convolutional Neural Networks Configurations Used}

Two CNNs architectures were used and compared, AlexNet \cite{NIPS2012_4824} and LeNet \cite{lecun1998gradient}. AlexNet is very good for image classification, first due to the depth of the architecture, and other factors like dropouts. Another important factor to note, AlexNet has been trained on 1000 classes with more than 1 million images, which implies that the weights are the depth of the network could be reused for fine-tuning. with some tweaking LeNet in its parameters performs better.

LeNet architecture takes as input grayscale images of size 28x28, whereas AlexNet takes as input color images of size 256x256. All images are normalized by subtracting the training mean from all the images before the training begins.
 
AlexNet has been slightly modified as follows: five convolutional layers, followed by max-pooling, norm, ReLU layers, and three fully-connected layers.

LeNet as well has been slightly modified as follows: two convolutional layers, followed by max-pooling layers and two inner products layers.

\begin{figure}[!t]
	\centering
	\includegraphics[scale=0.4]{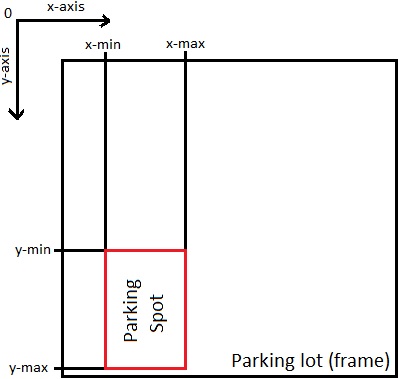}
	\caption{Definition of x-min, x-max, y-min, y-max of a single spot. Those coordinates are relative to the current image of the parking lot and will not change since the setup of the frame will remain static.}
	\label{fig:coordinates}
\end{figure}

\subsection{Processing and Prediction}

We show some details of the processing that happens in the system, from the image acquisition to the prediction of the specific status of the selected parking spot.

\subsubsection{Image Acquisition}

The images are taken from a video of a parking lot stored on a local workstation. Using OpenCV libraries, the video is read and generates frames to perform the predictions.

\subsubsection{Definition of Parking Spots}

The first frame of the footage (or a frame that contains mostly occupied parking spots) is used to define what a parking spot is in that current frame when the program starts. This technique of defining a parking spot on one frame relies on the fact that the camera will not move during the whole process of prediction and logically the parking lot does not move, only the cars are moving in or out of the different defined parking spots. Therefore this definition applies to the subsequent frames. The user defines the spot by clicking on two corners of the parking spot (the upper corner left and the lower corner right). After that set of clicks, the parking spot is defined and classified on the go.

\subsubsection{Storage of Coordinates on JSON file}

A JSON file is created by the system to hold the coordinates of the different parking spots defined by the user. The JSON coordinates (terminologies used in the JSON format are explicitly explained in \figurename{\ref{fig:coordinates}}) are stored in the following format:

\begin{verbatim}
{
"spot_1": {               /*Spot identifier*/
"name": "spot_1",       /*Spot identifier*/
"xmax": 448,  /*Spot maximum x-coordinate*/
"xmin": 345,  /*Spot minimum x-coordinate*/
"ymax": 542,  /*Spot maximum y-coordinate*/
"ymin": 478   /*Spot minimum y-coordinate*/
}, ... 
\end{verbatim}

\begin{figure}[!t]
	\begin{algorithm}[H]
		\caption{Classification}
		\begin{algorithmic}[1]
			\renewcommand{\algorithmicrequire}{\textbf{Input:}}
			\renewcommand{\algorithmicensure}{\textbf{Output:}}
			\REQUIRE model, deploy\_file, cropped\_image, label\_file
			\ENSURE  The actual class of the cropped image (\textbf{string})
			\STATE status = \{ \}; Dictionary holding each class and their corresponding confidence
			\STATE i = 0
			\STATE result\_status = " ", actual name of the class.
			\WHILE {all the labels and confidences of each predictions}
			\STATE status[confidence(i)] = confidence
			\STATE status[label(i)] = label
			\STATE i $\gets$ i + 1
			\ENDWHILE
			\IF {status[confidence0] $>$ status[confidence1]}
			\STATE result\_status = status[label0]
			\ELSE \STATE result\_status = status[label1]
			\ENDIF
			\RETURN result\_status 
		\end{algorithmic} 
		\label{classify}
	\end{algorithm}
\end{figure}

\subsubsection{Prediction and classification}

The JSON file previously generated is used to perform the classification of each spot. The following set of operations will take place:

\begin{enumerate}
	\item \textbf{Cropping}: OpenCV allows processing the spots on a frame-by-frame basis. The system takes as one of its parameter the JSON file that contains the set of coordinates and the video. Parking spots will be cropped from each frame of the video based on the JSON file. The cropping is done sequentially and the corresponding image is stored on the disk, in other words, only one cropped image is stored and the next picture overrides it later on.
	\item \textbf{Prediction and classification}: After the training phase, the generated Convolutional Neural Network model comes along with two other files: label.txt containing the different label of a spot (vacant or occupied) and the deploy.prototxt, which is the network configuration used by the model. This file defines the type of input allowed by the model and the protocol followed to produce the output. The system takes as arguments the model created at the training phase, the corresponding labels, the network configuration (deploy.prototxt), and the cropped image of the parking (one at a time). Those arguments are processed through a borrowed function written by Nvidia, classify\cite{nvidia_example}, which is a part of DiGITS. Modifications were made only in the classify function as shown in Algorithm \ref{classify}. Those modifications allow to visually determine whether a spot is either vacant (green rectangle) or occupied (red rectangle), based on the score or confidence level (result of the probability model computed) produced by the classification method. Those results are then streamed to the main system to produce the output.
	
\end{enumerate}

\begin{table}[!h]
	\begin{center}
		\caption{Configuration of the 2 datasets used for to train  the system}
		\label{table:data}
		\begin{tabular}{|c | c | c |c|} 
			\hline
			\textbf{Dataset} & \textbf{Dimensions} & \textbf{Size} & \textbf{Entries(train \& val)} \\ [0.3ex] 
			\hline
			Parking(Colour) & 256x256x3 & 84.3MB & 2026 \& 676 \\ 
			\hline
			Parking(Grayscale) & 28x28x1 & 2.16MB & 2026 \& 676 \\
			\hline
		\end{tabular}
	\end{center}
\end{table}

\section{Experiments and results}

\begin{figure}[!t]
	\centering
	\begin{subfigure}{0.4\textwidth}
		\centering
		\includegraphics[width=1\linewidth]{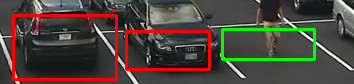}
		\caption{Prediction on some spots 3. The system does get triggered with a human passing through the classification area, which is classified as vacant. The spot 1 and 3 (first two rectangles to the left) are correctly classified are occupied.}
		\label{fig:free1busy}
	\end{subfigure}
	\hfill
	\begin{subfigure}{0.4\textwidth}
		\centering
		\includegraphics[width=1\linewidth]{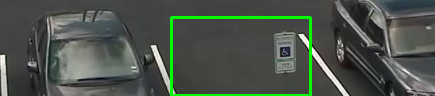}
		\caption{Prediction on some spots 2. The street sign does not trigger the classifier as it is too small to change the color distribution of the area covered by the green rectangle.}
		\label{fig:free1}
	\end{subfigure}
	\caption{System in action on predicting vacancy of parking spot}
	\label{fig:System}
\end{figure}

For the experiments, the comparison of LeNet and AlexNet allowed us to run the classification on a recorded video\footnote{https://www.youtube.com/watch?v=U7HRKjlXK-Y} containing a parking lot, with some modifications on the hardware and the weight of the inputs (images) that each network has to process.

The next section presents the different sets of CNN and parameters employed and the corresponding performances on the earlier stated video.

\subsection{Dataset: Input to the Convolutional Neural Networks}\label{data}

From Section \ref{thedataset}, a set of images of both statuses a parking spot: occupied or vacant. As shown in Table \ref{table:data}, we have a very small number of images in the dataset for training compared to the number of images in the dataset from Section \ref{thedataset}. Many instances of identical pictures were removed to avoid any type of overfitting. Based on the two selected CNNs to develop the system, we shaped the data according to the input requirements of the concerned CNN.

For the AlexNet, all the images are resized to 256x256. Those images are in RGB.

For the LeNet, all the images are resized to 28x28. Those images are in grayscale.

\begin{figure}[!t]
	\centering
	\includegraphics[scale=.4]{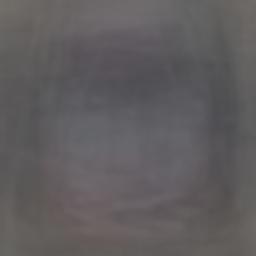}
	\caption{Image Mean of the dataset: Vacant and Occupied spots.}
	\label{fig:mean}
\end{figure}

From \figurename{\ref{fig:mean}}, the image mean is giving a clear approximation of all the status of a parking spot at any given time, including the parking lines.

Table \ref{table:data} shows the split of the dataset used for training and validation processes with the ratio of 75\% for training and 25\% for testing.

\subsection{Configurations of Convolutional Neural Networks}

Given the datasets generated from Section \ref{data}, we built some CNN configurations, based on LeNet and AlexNet, to use them in the system for classification. For all the network configurations, the Base Learning Rate or \textit{Base lr} is initialized at 0.01. The Learning Rate Policy or \textit{Policy lr} is the way the learning rate will change with time. The learning rate is be divided by 10 after each 33\% portion of the training set. Three solvers for the classifier are mainly used to compare the results. The three solvers used are:
\begin{enumerate}
	\item Stochastic Gradient Descent (SGD)
	\item Adaptive Gradient Descent (AdaGrad)
	\item Nesterov\textquotesingle s Accelerated Gradient (NAG)
\end{enumerate}

We have made 30 passes through the training set (epochs). Table \ref{table:data2} is the summary of all the custom configurations to get the best accuracy with the validation set along with the speed performance. The validation set is made out of 25\% of the entire set. The validation set is entirely independent of the training set.
\begin{table}[t!]
	\begin{center}
		\caption{All configurations for the system training using AlexNet and LeNet using different types of solvers}
		\label{table:data2}
		\begin{tabular}{|c | c | c | c |} 
			\hline
			\textbf{Network} & \textbf{Solver} & \textbf{Accuracy(Train)} & \textbf{Accuracy(Test)}\\ [0.5ex] 
			\hline
			AlexNet & SGD & 0.99 & 0.95 \\ 
			\hline
			AlexNet & NAG & 0.99 & Never Ended \\
			\hline
			AlexNet &  AdaGrad & 0.98 & 0.89 \\
			\hline
			LeNet & SGD & 0.99 & 0.92 \\ 
			\hline
			LeNet & NAG & 0.99 & 0.93 \\
			\hline
			LeNet & AdaGrad & 0.99 & 0.89 \\
			\hline
		\end{tabular}
	\end{center}
\end{table}

\subsection{Results}\label{results_experiements}

After the training phase, very high accuracy values are noted at validation level based on Table \ref{table:data2} (accuracy calculated during training with images that are closely similar to the training set, but not in the training set). For the accuracy of the system, PKLot dataset has been used \cite{deAlmeida20154937} to assess the strength of the system to classify on a foreign parking lot. 

Table \ref{table:data2} shows the testing results of the classifier after the training and validation process. The test data has been generated from unseen images of parking spots to the classifier. The performance of all different CNN configurations has been tested against the test images. All those configurations received the same batch of test images at once to simulate a close to real-time heavy flow. In the same table, during testing AlexNet with the Nesterov Accelerated Gradient, the batch of images took very long to terminate. Giving that the system is meant for a real-time purpose, time is an important factor to monitor and delays are considered as failure.

\figurename{\ref{fig:free1busy}} and \figurename{\ref{fig:free1}} represent the results produced by the system with AlexNet as the classifier. The change of lighting, which is a big factor misbehavior of many classifiers, does not affect too much the classification done by the system trained on this relatively small dataset. The classification runs on a single frame rather than a stream of frames, where instead of 25 frames per second (fps), it classifies on 5 fps, which takes into account the 2GB of the Nvidia graphic memory.

\section{Conclusion}
In this paper, we have shown with success that computer vision with a single camera, using Convolutional Neural Networks is having a success rate close to traditional methods used in the past. The classifier made of LeNet architecture with the Nesterov\textquotesingle s Accelerated Gradient as solver produced an accuracy of 93.64\%, and the AlexNet architecture with the Stochastic Gradient Descent as solver produced an accuracy of 95.49\% which are better than both Amato et al. \cite{7543901} and Wu et al. \cite{Wu} who produced respectively 90.4\% and 93.52\%. Given the hardware used for the research, the LeNet architecture with the Nesterov Accelerated Gradient is faster than the AlexNet architecture with the Stochastic Gradient Descent when tested.

The use of Convolutional Neural Networks facilitates the process of building a classifier as they automatically extract and use the features from the dataset. As shown in \figurename{\ref{fig:System}}, the system is able to produce an accurate output based on the situation of the current parking spot the camera is facing, given that spot is well defined by the user when launching the system. 

We have noticed during the research the presence of noise due to either the light, some elements stuck on the ground (oil leaks, cracks ...), or humans passing across parking spots. Those events have been corrected by improving the dataset based on the event of false classification due to noise. We have obtained the best combination in terms of parameters of the classifier by having the Nesterov Accelerated Gradient as solver and LeNet as network architecture, especially for the speed of execution compared to the better accuracy produced by the AlexNet architecture with the Stochastic Gradient, but with poor results when it comes to speed (speed is entirely relative to the hardware used). The classification on video is faster and accurate enough to be deployed in a real-life situation, on a parking that does not have any image appearing in the training set. It implies that the system can be used to a new parking without retraining of the classifier (given that the amount of noise is under control. E.g.: lighting, oil, or any type of leaks on the ground...).

For future work, we are planning on first improving the positioning of the camera so that we get a clear view of each car in the parking lot without a car hiding any part of the next car. Given that the availability of the parking spots is deduced from a well-defined parking lot, the system could direct the driver to the nearest available spot using some artificial intelligence techniques.


\section*{Acknowledgement}
I would like to thank the Council for Scientific and Industrial Research (CSIR), Adrian \cite{Adrian}, Prof. Turgay \c{C}elik, and the PIMD (Property and Infrastructure Management Division) at the University of the Witwatersrand for their contribution in this research project.



%

\bibliography{ref} 
\bibliographystyle{ieeetr}

\end{document}